\documentclass[letterpaper]{article} 
\usepackage{aaai2026}  
\usepackage{times}  
\usepackage{helvet}  
\usepackage{courier}  
\usepackage[hyphens]{url}  
\usepackage{graphicx} 
\usepackage{natbib}  
\usepackage{caption} 
\usepackage{algorithm}
\usepackage{algorithmic}
\usepackage{amsmath}
\usepackage{amssymb}
\usepackage{booktabs}
\newcommand{\xmark}{\text{\sffamily X}}

\usepackage{tikz}
\usepackage{newfloat}
\usepackage{listings}
\DeclareCaptionStyle{ruled}{labelfont=normalfont,labelsep=colon,strut=off} 
\floatstyle{ruled}
\newfloat{listing}{tb}{lst}{}
\floatname{listing}{Listing}
\title{Right Looks, Wrong Reasons: Compositional Fidelity in Text-to-Image Generation}
\author{
    Mayank Vatsa$^1$, Aparna Bharati$^2$, and Richa Singh$^1$
}
\affiliations{
    $^1$IIT Jodhpur, India,  \\$^2$Lehigh University, USA

    \{mvatsa, richa\}@iitj.ac.in, apb220@lehigh.edu
}

\usepackage{bibentry}

\begin{document}

\maketitle

\begin{abstract}

The architectural blueprint of today’s leading text-to-image models contains a fundamental flaw: an inability to handle logical composition. This survey investigates this breakdown across three core primitives—negation, counting, and spatial relations. Our analysis reveals a dramatic performance collapse: models that are accurate on single primitives fail precipitously when these are combined, exposing severe interference. We trace this failure to three key factors. First, training data show a near-total absence of explicit negations. Second, continuous attention architectures are fundamentally unsuitable for discrete logic. Third, evaluation metrics reward visual plausibility over constraint satisfaction. By analyzing recent benchmarks and methods, we show that current solutions and simple scaling cannot bridge this gap. Achieving genuine compositionality, we conclude, will require fundamental advances in representation and reasoning rather than incremental adjustments to existing architectures.
\end{abstract}

\section{Introduction}

\begin{figure}[t]
\centering
\includegraphics[width=0.965\columnwidth]{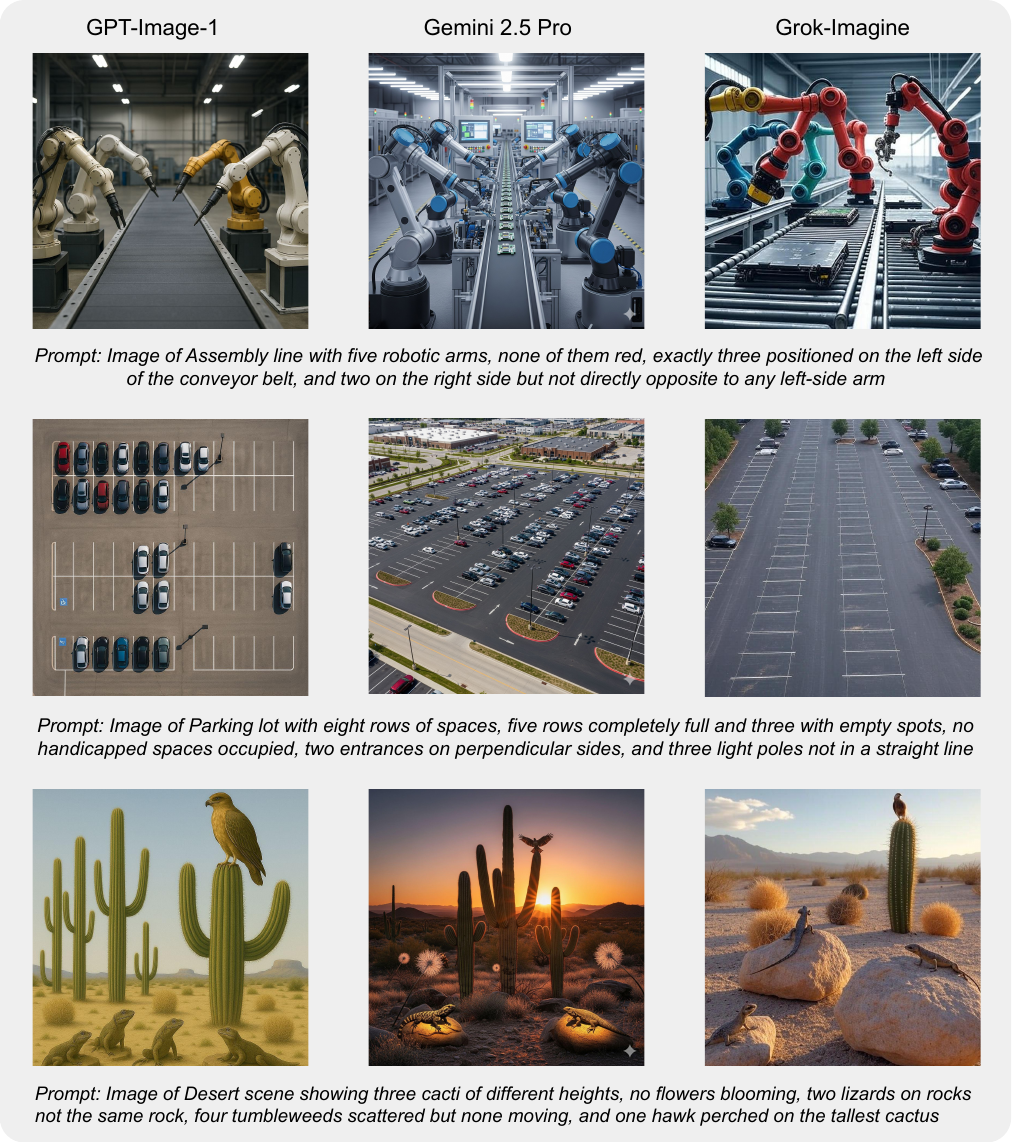} 
\caption{Examples of state-of-the-art Text-to-Image Generation models failing to correctly bind attributes and satisfy spatial relations in complex prompts. 
}
    \label{fig:taxonomy}
    \vspace{-12pt}
\end{figure}

Text-to-image (T2I) generation systems such as DALL$\cdot$E~3~\cite{Betker2023Dalle3}, Stable Diffusion~\cite{Rombach2022HighResolutionIS}, Imagen~\cite{Saharia2022PhotorealisticTD}, and Parti~\cite{Yu2022ScalingAM} can produce remarkably photorealistic images that capture style, aesthetics, and individual concepts with impressive fidelity. Despite their visual sophistication, these models systematically fail at compositional reasoning, i.e., the ability to satisfy multiple constraints such as counting, attribute binding, spatial relations, and negation, simultaneously. For example, the prompt “exactly three red apples to the left of a vase with no flowers” is trivial for humans but challenging for these models. As illustrated in Figure \ref{fig:taxonomy}, even when a model can satisfy constraints individually, combining them causes the performance to drop significantly~\cite{Huang2023T2ICompBench,Hsieh2023SugarCrepe,li2025enhancing, s11263-025-02371-0}.

The failure modes of these models are now well documented. For negation, models often ignore cues and permit forbidden content~\cite{Alhamoud2025VLMNegation,Singh2025CCNeg,Patel2024TripletCLIP}. For counting, errors grow with target cardinality, and recent work proposes targeted remedies within T2I pipelines~\cite{Paiss2023TeachingCLIP,Binyamin2025CountGen,Zafar2024DetectionDriven}. For spatial relations, systems may satisfy local or pairwise predicates while violating global layout consistency~\cite{Gokhale2023BenchmarkingSpatial,Kamath2023WhatsUp,Cho2023DSG}. These failures stem from architectural biases such as continuous attention, which poorly handles discrete logic, and data sparsity, as web-scale captions rarely cover structured, negated, or high-count cases.

These shortcomings carry real-world consequences. Educational platforms cannot reliably render illustrations with specific counts. Technical documentation tools misplace components in assembly diagrams. Under too many constraints, scientific and medical illustrations can show anatomically or physically impossible results.
Although extensive iteration can lead to intended compositions, this undermines direct language control and hinders AI's potential in high-stakes domains. Compositional fidelity is thus a prerequisite for reliable and trustworthy deployment \cite{mittal2024responsible}.

This survey analyzes \emph{compositional fidelity} through the lens of \emph{compositional primitives}, defined as the semantic mechanisms that combine to specify a scene. These include attribute binding; size and scale relations, actions and temporal order, part–whole structure, logical connectives (AND/OR/XOR), quantifier scope and coreference, symmetry and topology beyond pairwise relations, material/texture binding, occlusion reasoning, and 3D/viewpoint consistency~\cite{Thrush2022Winoground,Ma2023CREPE}. We focus on three fundamental primitives: negation (absence), counting (cardinality), and spatial relations (geometric/topological layout). 

This paper has three key contributions. (1) We provide a formal account of the intersectional failure, explaining why progress on isolated capabilities does not extend to joint prompts, and connect them to constrained optimization and combinatorial hardness. (2) We synthesize methods: data augmentation and contrastive training for negation; architectural strategies and mixture-of-experts for counting; layout and structural control for spatial relations; and hybrid neural-symbolic pipelines for joint composition. (3) We review 15 benchmarks, showcasing a shift from human studies to automated and adversarial evaluations, and discuss strengths, biases, and gaps. Finally, we propose a research direction that spans theory, architecture, training, and evaluation to bridge visual plausibility and logical faithfulness\footnote{While prior surveys center on architectures, scaling trends, or benchmarks~\cite{Thrush2022Winoground,Ma2023CREPE,Vempati2023VALSEAV,Yuksekgonul2023ARO}, we focus on compositional fidelity, introducing a primitive-based taxonomy that explains why scale alone falls short and points to promising solutions.}.

\section{Problem Formulation}
\label{sec:problem}

Let $\mathcal{X} = [0,1]^{H\times W\times 3}$ denote the RGB image space where $H$ and $W$ are the image height and width, and $\mathcal{Y}$ the space of natural-language prompts.
A T2I model $G_\theta:\mathcal{Y}\!\to\!\Delta(\mathcal{X})$ induces the conditional distribution $p_\theta(x|y)$ for $y\!\in\!\mathcal{Y}$, where $\Delta(\mathcal{X})$ denotes the space of probability distributions over $\mathcal{X}$.
Modern systems, including diffusion-based (e.g., Stable Diffusion~\cite{Rombach2022HighResolutionIS}), autoregressive (e.g., Parti~\cite{Yu2022ScalingAM}), and proprietary systems (e.g., DALL$\cdot$E~3~\cite{Betker2023Dalle3}), map linguistic concepts to visual content using massive paired corpora.
Despite impressive perceptual quality, early probes documented \emph{compositional} failures on instructions that require logical structure and grounding~\cite{Saharia2022PhotorealisticTD}.
Similar compositional challenges have been observed in vision-language models like CLIP~\cite{Radford2021LearningTV}, which struggle with relational semantics despite strong object-level understanding.

\subsection{Definition \& Significance of Primitives in T2I Gen.}



\subsubsection{Negation - The Challenge of Concept Absence:}
Negation requires reasoning about what should \emph{not} appear (e.g., ``a bowl of fruit \emph{without} bananas'').
Let $y_{\mathrm{aff}}$ be an affirmative prompt and $y_{\mathrm{neg}}$ its negated counterpart. With latent $z$, negated prompts plausibly span more admissible scenes~\cite{Vempati2023VALSEAV,Patel2024TripletCLIP,Alhamoud2025VLMNegation},
$\mathcal{H}\!\big(p(z\mid y_{\mathrm{neg}})\big)\;>\;\mathcal{H}\!\big(p(z\mid y_{\mathrm{aff}})\big)$
where $\mathcal{H}(\cdot)$ denotes the Shannon entropy. In T2I prompts, negation appears in a range of recurring expressions that operate at different semantic levels. \textbf{Morphological} negation uses affixes (e.g., \emph{un}striped, \emph{non}toxic, sugar\emph{less}) to flip attributes. \textbf{Lexical/privative} adjectives (e.g., \emph{empty}, \emph{barefoot}, \emph{devoid of}) express the absence of object or content. \textbf{Syntactic/clausal} cues (e.g., \emph{no}, \emph{not}, \emph{without}, \emph{neither...nor}) deny entity presence or relations. \textbf{Quantified} negation (e.g., \emph{no} $N$, \emph{fewer than} $n$) sets zero or upper-bound constraints on counts. \textbf{Relational} negation (e.g., \emph{not left of}, \emph{not touching}) denies specific spatial predicates.

\noindent\textbf{Why this matters in T2I:}
Negation bridges explicit linguistic markers and implicit visual signals of absence—a mapping current models handle poorly. It is rare in web captions~\cite{Truong2023Naysayers, Hossain2020NegationMT} and competes with strong co-occurrence priors. In practice, we observe recurring failure patterns~\cite{Alhamoud2025VLMNegation}. Models ignore negation markers, allowing forbidden objects to appear. Scope is misapplied, with ``not'' attaching to the wrong phrase. Models over-suppress content as a shortcut. UI-level ``negative prompts'' alter text guidance but do not guarantee semantic absence. Occlusion confounds evaluation, as hidden items can be mistaken for true absence or presence by automated judges. These issues amplify on joint prompts like ``\emph{exactly two} red apples \emph{not touching}'', which require enforcing multiple types of absence while meeting precise positive constraints. In high-stakes domains (e.g., medical: ``\emph{no} fracture''), such failures yield misleading imagery.

\subsubsection{Counting - Precise Cardinality Control:}
Counting, which demands precise control over the number of instances (e.g., ``exactly five roses''), exposes a core architectural limitation of modern generators. Because Transformer attention operates in parallel, it lacks an explicit enumerator, forcing numerosity to be encoded implicitly within continuous weights rather than through discrete counting steps. This fundamental mismatch between the task and the architecture leads to a predictable pattern of failure. Empirically, accuracy degrades as the target number $n$ increases, with errors compounding in cluttered or occluded scenes~\cite{Huang2023T2ICompBench,Paiss2023TeachingCLIP,Zafar2024DetectionDriven,Binyamin2025CountGen}. Rather than a gradual decline, this degradation is super-linear: error growth consistently fits a power law with exponent $\beta \in [1.2,1.5]$ across multiple benchmarks,
\begin{equation}
\label{eq:counting-error}
\mathrm{Error}(n) \;\approx\; \Theta\!\big(n^{\beta}\big), \quad \beta \in [1.2, 1.5].
\end{equation}




Counting in T2I prompts manifests in a variety of complex linguistic forms. Models must handle \emph{exact numerals} (``exactly three red apples''), interpret \emph{bounded ranges} (``between three and five''), and resolve \emph{vague quantifiers} (``few,'' ``several''). The task can also be relational, involving \emph{comparatives} (``more apples than pears''), or compound, requiring \emph{multi-type specifications} (``three apples and two pears''). Furthermore, counts are often distributed across the scene via \emph{spatially partitioned tallies} (``two vases on the left, one on the right''). Fulfilling these prompts requires the model to successfully individuate instances, bind their attributes, and maintain count coherence across the entire composition.


\noindent\textbf{Why this matters in T2I:}
Counting exposes the limits of discrete reasoning in architectures that operate on continuous representations. We observe consistent failure patterns~\cite{Kirstain2023PickAPic, Binyamin2025CountGen}
such as objects accidentally duplicated or merged under clutter. Attribute-binding leakage causes numbers to apply to wrong subsets. Comparative constraints fail when one category disappears rather than adjusting counts appropriately. These issues compound with other primitives. A prompt like ``\emph{exactly three} red apples \emph{to the left of} a vase \emph{with no flowers}'' demands precise counting with correct attribute binding, specific spatial arrangement, and enforced absence of both extra apples and flowers. Any degradation in individuation, binding, or layout violates the instruction, making counting central to compositional fidelity.

\subsubsection{Spatial Relations - Geometric \& Topological Reasoning:}
Spatial relations dictate geometric object arrangements. A prompt like ``a blue cube \emph{on} a red sphere'' requires parsing linguistic relations into geometrically consistent scenes~\cite{Gokhale2023BenchmarkingSpatial,Kamath2023WhatsUp} and can be decomposed as:
\begin{multline}
\label{eq:logical_decomp}
\exists\, c, s \ \text{s.t.}\ \mathrm{IsCube}(c)\ \land\ \mathrm{IsSphere}(s)\ \land\ \mathrm{Color}(c,\text{blue})\\
\land\ \mathrm{Color}(s,\text{red})\ \land\ \mathrm{On}(c,s).
\end{multline}
While models often satisfy unary predicates (types, colors), enforcing the relational predicate \(\mathrm{On}(c,s)\) proves notably harder. In a viewer-centric frame, let object placements be axis-aligned rectangles \(b_c=(x_c,y_c,w_c,h_c)\) and \(b_s=(x_s,y_s,w_s,h_s)\). 
\(\mathrm{On}(c,s)\) requires the following:
\begin{equation}
\label{eq:geometric_constraint}
\begin{aligned}
& |y_c + h_c - y_s| < \epsilon, \\
& [x_c,\,x_c+w_c]\ \cap\ [x_s,\,x_s+w_s]\ \neq\ \emptyset,
\end{aligned}
\end{equation}
Here, ($x,y,w,h$) denote the top-left position and size of each axis-aligned bounding box, and $\epsilon$ is a proximity threshold. The first constraint ensures the bottom of \(c\) is near the top of \(s\), while the second ensures horizontal overlap. Object- or world-centric frames yield different but analogous conditions.

To generate coherent scenes, models must deconstruct the complex language of \textit{spatiality}, which manifests in multiple forms~\cite{Liu2023VisualSpatial}. The process begins with parsing fundamental directional relations that specify relative positions, such as ``left of'' or ``above.'' Building on this, topological relations encode concepts of containment and overlap, like ``in'' or ``on.'' The challenge becomes more fine-grained with proximity relations that indicate distance (``near,'' ``touching'') and alignment patterns that create structured arrangements (``in a row''). Furthermore, models must grasp physical plausibility through \emph{support} relations (``on top of'') and manage complex distributions with partitioned layouts (``two vases on the left, one on the right''). For clarity, we assume a viewer-centric interpretation for these relations unless a prompt specifies otherwise.


\noindent\textbf{Why this matters in T2I:}
Spatial reasoning connects textual semantics with geometric feasibility. Current models exhibit characteristic failures~\cite{Vempati2023VALSEAV, zhang2024compassenhancingspatialunderstanding}. Pairwise predicates hold locally but conflict globally. Directional relations flip under viewpoint changes. Support relations are nominally satisfied while violating physical plausibility. Contact constraints fail due to slight boundary overlaps. Structured patterns drift under clutter. These errors intensify on joint prompts like ``\emph{exactly three} red apples \emph{to the left of} a vase \emph{not touching} it'', which simultaneously require precise counting, correct attribute binding, specific layout, and absence of forbidden contacts. In high-stakes settings (e.g., medical: ``lesion \emph{left of} the hippocampus''), mis-grounded spatial relations yield visually credible but semantically incorrect images.

\paragraph{Joint Compositionality - The Compounding Challenge:}
The most significant failures occur when primitives are combined. Consider ``two blue cubes to the left of a vase \emph{with no flowers}''—a prompt joining counting, spatial relations, attribute binding, and negation. Failure rates on such prompts are higher than single primitives~\cite{Huang2023T2ICompBench}. 

Let $F_\theta(y)$ denote compositional faithfulness: the probability that all constraints in prompt $y$ are satisfied. 
Under independence of counting, spatial, and negation, joint success equals the product of individual rates.
\begin{equation}
\label{eq:ind-baseline}
F_\theta^{\mathrm{ind}}(y)\;:=\;F_\theta^{\mathrm{cnt}}(y)\,F_\theta^{\mathrm{spat}}(y)\,F_\theta^{\mathrm{neg}}(y).
\end{equation}
However, we consistently observe \emph{submultiplicative} performance~\cite{Huang2023T2ICompBench}. We quantify interference via:
\begin{equation}
\label{eq:submult}
\rho(y)\;:=\;\frac{F_\theta(y)}{F_\theta^{\mathrm{ind}}(y)}\
\end{equation}
$\rho(y)=1$ indicates independence and $\rho(y)<1$ indicates intersectional failures. Equivalently, the additive gap $\Delta_{\mathrm{joint}}(y)=F_\theta^{\mathrm{ind}}(y)-F_\theta(y)>0$ measures the deficit. We define interference as \emph{pronounced} when $\rho(y)\le \tau_\rho$ for threshold $\tau_\rho\in(0,1)$ (e.g., $0.75$). To illustrate the compounding nature, consider a scenario where individual primitives achieve moderate success rates (e.g., 70\% each). Under independence, joint faithfulness would be $0.7^3 = 34.3\%$. However, due to interference ($\rho(y) \approx 0.58$), actual performance drops to $\approx 20\%$, highlighting the challenge.

\noindent\textbf{Why this matters in T2I:}
Joint prompts expose hidden interactions between capabilities. Models exhibit constraint trading: enforcing layout breaks counting; honoring negation removes desired context; correct counts appear with incorrect attribute binding; and pairwise relations hold locally while global arrangement fails \cite{Kamath2024HardPositives}. These behaviors reflect architectural limitations such as continuous attention without explicit enumeration and training biases such as sparse negation exposure and strong co-occurrence priors~\cite{Yuksekgonul2023BoW}. The submultiplicative gap ($\rho(y)<1$) separates surface realism from genuine compositional reasoning, motivating targeted methods like inference-time composition~\cite{Feng2023ComposableDiffusion} and layout-guided image generation~\cite{Zheng2023LayoutDiffusion}.

\begin{table*}[]
\centering
\scriptsize
\begin{tabular}{llccccl}
\toprule
\textbf{Benchmark} & \textbf{Size} & \textbf{Auto} & \textbf{Neg} & \textbf{Count} & \textbf{Spatial} & \textbf{Primary Focus} \\
\midrule
DrawBench~\cite{Saharia2022PhotorealisticTD} & 200 & \xmark & \xmark & \checkmark & \checkmark & Human-eval comprehensiveness \\
PartiPrompts~\cite{Yu2022ScalingAM} & 1{,}600+ & \xmark & \checkmark & \checkmark & \checkmark & Multi-aspect evaluation \\
Winoground~\cite{Thrush2022Winoground} & 400 ex.\ (1{,}600 pairs) & \checkmark & \xmark & \xmark & \checkmark & Visio-linguistic composition \\
CREPE~\cite{Ma2023CREPE} & 370K+ & \checkmark & \checkmark & \xmark & \checkmark & Systematic compositionality \\
VALSE~\cite{Vempati2023VALSEAV} & $\sim$8{,}782 & \checkmark & \checkmark & \checkmark & \checkmark & Linguistic variations \\
ARO~\cite{Yuksekgonul2023ARO} & 50K+ & \checkmark & \xmark & \xmark & \checkmark & Attribution, relation, order \\
POPE~\cite{Li2023POPE} & variable & \checkmark & \checkmark & \xmark & \xmark & Object hallucination \\
T2I-CompBench~\cite{Huang2023T2ICompBench} & 6{,}000 & \checkmark & \checkmark & \checkmark & \checkmark & Comprehensive automated eval \\
DSG~\cite{Cho2023DSG} & 1{,}060 & \checkmark & \xmark & \checkmark & \checkmark & Scene-graph grounding \\
TIFA~\cite{10377168} & $\sim$4{,}000 & \checkmark & \checkmark & \checkmark & \checkmark & VQA-based faithfulness \\
PhysiComp~\cite{Chen2024PhysiComp} & $\sim$1K & \checkmark & \xmark & \checkmark & \checkmark & Physical plausibility \\
SugarCrepe~\cite{Hsieh2023SugarCrepe} & $>$1{,}000 & \checkmark & \checkmark & \checkmark & \checkmark & Hard negative probes \\
CC\text{-}Neg~\cite{Singh2025CCNeg} & 228K & \checkmark & \checkmark & \xmark & \xmark & Negation training/eval \\
NegBench~\cite{Alhamoud2025VLMNegation} & 79K & \checkmark & \checkmark & \xmark & \xmark & Comprehensive negation \\
CoCoCount~\cite{Binyamin2025CountGen} & 200 & \checkmark & \xmark & \checkmark & \xmark & Counting specialization \\
\bottomrule
\end{tabular}
\caption{Comparing key compositional benchmarks for T2I evaluation. \textbf{Auto}: supports automated evaluation. \textbf{Neg}/\textbf{Count}/\textbf{Spatial}: explicitly tests negation/counting/spatial relations. Sizes are from original papers or official repos; ``variable'' indicates no fixed canonical size.}
\label{tab:benchmarks}
\vspace{-10pt}
\end{table*}

\section{Literature Review}
\label{sec:related}

Having established the challenges associated with the three compositional primitives, we now survey the benchmarks and methods proposed to measure and address these failures. 

\noindent\textbf{Methods for Addressing Negation:} 
Early work traced negation blindness in T2I to data sparsity and architectural bias~\cite{Hossain2020NegationMT,Kassner2020NegatedMisprimed,Truong2023Naysayers}, and mainly proposed three mitigation avenues. Firstly, contrastive strategies such as TripletCLIP which builds triplets differing only by negation markers
and NegCLIP which generates adversarial negatives automatically~\cite{Yuksekgonul2023ARO} show promise but struggle with complex negation. 
Secondly, data augmentation strategies address the scarcity directly~\cite{Brooks2023InstructPix2Pix}. CC-Neg~\cite{Singh2025CCNeg} dataset contains 228K image-caption pairs 
that cover morphological, syntactic, and semantic negation across diverse visual contexts and helps in improving negation understanding.
Beyond data, architectural methods preserve negation by reducing word loss with asymmetric visual-semantic embeddings~\cite{Liu2025AVSE}, modeling it as an energy constraint~\cite{Feng2023ComposableDiffusion}, or encoding absence with empty boxes~\cite{Li2023GLIGEN}.
These structural approaches generally outperform data-driven contrastive methods, which struggle with complex negation and are prone to bias or overfitting from synthetic augmentations~\cite{Hsieh2023SugarCrepe}.

\noindent\textbf{Approaches to Counting:} Efforts to improve counting in T2I models follow two main paths. The first is data-centric: DALL·E 3, for instance, focuses on enhancing caption quality to ensure training data contains precise numerical information~\cite{Betker2023Dalle3}. The second path involves architectural innovations. These include modifying the attention mechanism, as seen in Stable Diffusion 3 and SDXL~\cite{Esser2024SD3,Podell2023SDXL,Kang2025CountingGuidance}, using a mixture-of-experts to increase numeric capacity~\cite{Xue2023RAPHAEL}, or applying bounded attention to prevent the merging of distinct objects~\cite{Dahary2024BoundedAttention}. However, the success of these is often limited to a small number of objects.


To enhance counting in T2I generation, layout-based methods enforce precise counts via spatial conditioning, using bounding boxes~\cite{Li2023GLIGEN}, semantic regions~\cite{Yang2023ReCo}, or structural maps~\cite{Zhang2023ControlNet}. These, however, limit the flexibility of text-only prompts. Hybrid approaches, like LLM-grounded diffusion, restore this by parsing text into structured scene representations, reducing constraint interference ($\rho(y) < 1$)~\cite{Lian2023LLMGroundedDiffusion}. Dynamic methods, such as RPG’s iterative refinement with multimodal feedback~\cite{Yang2024RPG}, balance counting with other constraints, while Visual Programming decomposes tasks into modular steps~\cite{Gupta2023VisualProgramming}. Alternatively, CountGen embeds a differentiable counting loss in training~\cite{Binyamin2025CountGen}, and detection-driven methods iteratively refine outputs to ensure accurate counts~\cite{Zafar2024DetectionDriven}.

\noindent\textbf{Spatial Relation Methods:}  
Spatial reasoning improvements ground language in geometric control. Many approaches use layout-based methods for direct spatial conditioning. Some methods integrate this control during training. GLIGEN, for instance, uses box conditioning to guide generation~\cite{Li2023GLIGEN}. ControlNet leverages structural cues like edges and keypoints~\cite{Zhang2023ControlNet}. LayoutDiffusion also employs semantic segmentation for object adjacencies~\cite{Zheng2023LayoutDiffusion}. Other methods offer training-free control by manipulating attention at inference time. ZestGuide, for example, achieves zero-shot control via attention manipulation~\cite{CouaironICCV2023}. Similarly, attention refocusing optimizes layout supervision during the sampling process \cite{phung2023}.

Approaches to spatial grounding can be broadly categorized into sampling-based, attention-based, architectural, and 3D-aware strategies. Sampling-based methods include Composable Diffusion, which decomposes spatial relations into separate energy functions that combine during generation~\cite{Feng2023ComposableDiffusion}, and MultiDiffusion, which enforces scene coherence by generating multiple aligned views that are fused into a consistent whole~\cite{BarTal2023MultiDiffusion}. Attention-based methods such as Attend-and-Excite refine attention maps iteratively until spatial constraints are satisfied~\cite{Chefer2023AttendExcite,Rassin2023LinguisticBinding}. Architectural approaches embed spatial reasoning more directly: Set-of-Mark prompting uses visual markers to disambiguate references~\cite{Yang2023SetOfMark}, while CoMPaSS curates accurate training pairs with spatial constraints and strengthens text encoders through ordered token encoding~\cite{zhang2024compassenhancingspatialunderstanding}. Finally, 3D-aware methods such as Zero123 learn viewpoint-consistent representations, ensuring coherence across multiple perspectives~\cite{Liu2023Zero123}.
However, layout-based methods often require additional structural inputs that may not scale to complex scenes.

\noindent\textbf{Joint Compositional Methods:} To address the performance collapse when multiple primitives are combined, specialized approaches have been developed to mitigate interference. One strategy is inference-time composition, which processes different constraints separately to prevent them from conflicting with one another during generation~\cite{Feng2023ComposableDiffusion, Lian2023LLMGroundedDiffusion, Xue2023RAPHAEL}. Alternatively, training-based approaches target compositional generalization. For instance, compositional augmentation integrates primitives systematically in training data to enhance joint handling~\cite{li2025enhancing, s11263-025-02371-0}. On the other hand, EvoGen leverages curriculum learning with language models and visual checkers to progressively refine compositional skills~\cite{han2025progressive}. Despite their promise, these training strategies often face challenges. Catastrophic forgetting and distribution shift often arise, where enhancing one primitive via augmentation can degrade others due to overfitting~\cite{Hsieh2023SugarCrepe}.


\noindent\textbf{Evaluation Benchmarks:} Progress in compositional generation has been enabled by increasingly sophisticated evaluation frameworks. Table~\ref{tab:benchmarks} summarizes fifteen major benchmarks spanning 2022-2025. Early human-evaluated benchmarks like DrawBench~\cite{Saharia2022PhotorealisticTD} and PartiPrompts~\cite{Yu2022ScalingAM} established the problem space but faced scalability limitations. The shift to automated evaluation enabled systematic assessment at scale, with T2I-CompBench evaluating prompts across all three primitives~\cite{Huang2023T2ICompBench} and CREPE providing 370K+ variations~\cite{Ma2023CREPE}. More recent benchmarks emphasize adversarial testing through hard negatives and VQA-based probing. While existing frameworks cover negation, counting, and spatial relations, critical gaps remain. Key missing areas include temporal reasoning and complex multi-object interactions. Furthermore, assessing physical plausibility requires moving beyond current specialized efforts.



\section{Synthesis and Open Challenges}
\label{sec:synthesis}

Our analysis of negation, counting, and spatial relations highlights fundamental limitations in current text-to-image generation systems. The submultiplicative performance degradation quantified in Equation~\ref{eq:submult} emerges from deeper architectural and methodological constraints. We summarize these findings into actionable insights and present future directions toward compositionally faithful generation.

\noindent\textbf{The Data-Architecture Mismatch:} The compositional failures we observe stem from a fundamental misalignment between training data distributions and architectural inductive biases. Consider the joint distribution of compositional primitives in training data:
\begin{equation}
p_{\text{train}}(\text{neg}, \text{count}, \text{spat.}) \;\approx\; p(\text{neg}) \cdot p(\text{count}) \cdot p(\text{spat.}),
\end{equation}
an independence approximation that highlights the extreme sparsity of joint cases. In reality, not only are the marginal probabilities of each primitive low, but their co-occurrence is even rarer, with some pairs exhibiting negative correlations. For example, in popular datasets, captions with explicit negation are rare: MS~COCO ($\sim0.4\%$), CC3M ($1.63\%$), CC12M ($\sim2.5\%$), and LAION--400M ($\sim0.6\%$); larger corpora such as LAION--2B and DataComp--1B show similar low sub-percent to low single-digit rates depending on cues like \emph{no/not/without}~\cite{Bui2024NeIn,Park2025KnowNoBetter}. High-count scenes ($n>5$) appear in under $2\%$ of samples, and complex spatial arrangements with multiple relations constitute less than $5\%$. Thus, the effective probability of encountering prompts that combine negation, counting, and spatial constraints during training is vanishingly small.



The architectural response to this sparsity is predictable but problematic. Continuous attention mechanisms learn to approximate the majority modes:
\begin{align}
\label{eq:scoring}
x^{\ast}(y) &= \arg\max_{x \in X} \; S_\theta(x \mid y), \nonumber \\
S_\theta(x \mid y) &= \sum_{i} \alpha_i \, f_i(x,y) + \lambda R(x).
\end{align}
where $S_\theta(x|y)$ denotes the composite score over constraint-specific functions $f_i(x,y)$ with learned weights $\alpha_i$, and $R(x)$ encodes regularization or visual (realism) priors. The regularization term dominates when compositional requirements conflict with learned priors, explaining why models generate plausible yet unfaithful images. This mismatch manifests differently across primitives. Models lack explicit mechanisms to represent negation:
\begin{equation}
\mathcal{L}_{\text{neg}} = \min_\theta \mathbb{E}_{y \sim p(y)} \left[ D(G_\theta(y), \emptyset_{\text{target}}) \right]
\end{equation}
where $\emptyset_{\text{target}}$ denotes the absence of specific content and $D$ is a distance/divergence measure. Current architectures cannot differentiate this from general content suppression, leading to over-deletion or complete ignorance.

\noindent\textbf{Emergent Complexity in Joint Composition:}
The submultiplicative performance reflects genuine computational complexity rather than simple capability multiplication. When primitives combine, the constraint satisfaction problem becomes NP-hard in the general case\footnote{Satisfying sets of qualitative spatial constraints is NP-complete in general (RCC-8);
the analogous temporal algebra is NP-hard/NP-complete; and related layout problems
such as cartographic label placement are NP-hard, highlighting the combinatorial
nature of global spatial composition~\cite{Renz1999RCC8,Nebel1995Allen,Formann1991MapLabeling}.}. Consider a prompt with $n$ objects, $m$ spatial relations, and $k$ negation constraints. The search space grows as: $|\mathcal{S}| = O(n! \cdot 2^m \cdot \binom{n}{k})$. Current models use greedy local search, leading to characteristic failures: objects meet local constraints but break global consistency, counts hold until spatial layout is enforced, and negation applies to the wrong scope when combined with relations. We can formalize the joint generation problem as a constrained optimization:
$x^* = \arg\max_{x \in \mathcal{X}} p_\theta(x|y)$
such that $C_{\text{count}}(x, y) = 1$, $C_{\text{spatial}}(x, y) = 1$, and $C_{\text{neg}}(x, y) = 1$, where $C_i$ are binary constraint satisfaction functions. Current methods only partially satisfy these constraints, leading to compositional failures.

\noindent\textbf{Inter-Primitive Interference:}
The submultiplicative performance ($\rho(y) < 1$) arises from interference between primitives, exacerbating issues like hallucinations and physical implausibility. For instance, enforcing negation (e.g., ``no flowers'') may trigger hallucinations of unrelated objects due to strong co-occurrence priors in training data~\cite{Li2023POPE}. Similarly, combining counting and spatial constraints can produce physically implausible scenes, such as objects defying gravity when satisfying local relations~\cite{Chen2024PhysiComp}. These interactions highlight the need for models to disentangle primitives explicitly, through structured representations or constraint prioritization, to mitigate unintended outputs while preserving compositional fidelity.

\subsection{Critical Research Gaps}

\noindent\textbf{Theoretical Foundations:} The theory for compositional generation remains largely unwritten and critical questions unanswered. For instance, the computational lower bounds for faithful composition are unknown, and we cannot yet formally characterize which compositional patterns are learnable from the data. Mechanistic interpretability has shown that attention heads specialize in different primitives (negation and counting)~\cite{Rassin2023LinguisticBinding, Yuksekgonul2023BoW}. However, the interaction mechanisms between these specialized components remain largely opaque to current analytical methods. Empirical evidence already suggests that merely increasing model scale is not a viable solution; larger models show marginal gains on isolated primitives but suffer the same performance collapse on joint tasks~\cite{Esser2024SD3, Betker2023Dalle3}. This points to a deeper architectural limitation. Exploring connections to classical constraint satisfaction, such as SAT solving or neuro-symbolic methods~\cite{mao2018the}, could provide a path forward. However, these parallels are currently underexplored, and formalizing them could inspire the architectural innovations needed to overcome today's compositional failures.



\noindent\textbf{Evaluation Methodology:}
Current evaluation paradigms reveal significant limitations across the literature. Human evaluation provides nuanced judgment but suffers from inconsistency and lacks scalability~\cite{Saharia2022PhotorealisticTD}. Automated metrics using object detectors and VQA models introduce systematic biases~\cite{10377168}. The correlation between existing metrics and human compositional assessment remains weak, particularly for complex multi-constraint prompts~\cite{Ghosh2023GENEVAL}. 
Natural language ambiguity makes ground truth specification difficult—``a few cats near some boxes'' admits multiple valid interpretations. Occlusion and viewpoint variation further complicate accurate counting and spatial assessment. The trade-off between aesthetic quality and compositional fidelity is poorly characterized, with humans often preferring visually appealing but compositionally incorrect images.  
Evaluation frameworks that measure compositional generalization on novel combinations are an open challenge.

\noindent\textbf{Architectural Innovations:} Transformers excel at semantic understanding but lack explicit mechanisms for logical operations. Recent neuro-symbolic approaches show promise~\cite{Lian2023LLMGroundedDiffusion}, yet integration remains empirically driven rather than principled. 
Architectural directions such as modular architectures that decouple reasoning from generation, memory-augmented networks for explicit state, and hierarchical models separating “what,” “where,” and “how many”, merit deeper investigations. Intermediate representations such as scene graphs and layouts improve fidelity~\cite{Li2023GLIGEN} but depend on costly annotations, leaving the challenge of learning structured representations from text alone unresolved~\cite{Feng2023ComposableDiffusion}.

\noindent\textbf{Training Paradigms:}
Current training objectives, such as reconstruction for autoregressive models and denoising for diffusion models, optimize average-case performance on common patterns. These objectives misalign with compositional requirements that demand worst-case guarantees on logical constraints. The literature lacks frameworks explicitly designed for compositional fidelity. Curriculum learning with progressive complexity helps specific primitives but requires careful engineering~\cite{mao2018the, 10208670}. Compositional data augmentation improves targeted capabilities but struggles with diverse or out-of-distribution prompts~\cite{Hsieh2023SugarCrepe}. Reinforcement learning from human feedback could align models with compositional objectives; however, reward specification for complex constraints is an open problem~\cite{Li2025AlignBind}. In addition, data efficiency questions persist in the literature. 
Synthetic data risk distribution shift and massive augmentation yields only marginal gains, suggesting fundamental inefficiency in learning compositional patterns.




\noindent\textbf{Future Directions:} The path forward requires coordinated advances across multiple fronts. Better benchmarks must measure compositional generalization rather than memorization. Architectural innovations should balance continuous and discrete reasoning \cite{Thakral_2025_CVPR, 
dosi2025harmonizing}. Training paradigms must align logical objectives with statistical likelihood, and theoretical frameworks should guide development beyond empirical trial-and-error. The submultiplicative performance gap remains a critical metric of progress, and closing it demands fundamental advances. As the field matures, compositional fidelity becomes not just a technical challenge but a prerequisite for practical deployment. Techniques developed for negation, counting, and spatial relations will extend to temporal reasoning, causal relations, and abstract concepts, paving the way for controllable, reliable, and truly intelligent image generation.


\section{Conclusion}
This survey examined compositional fidelity in text-to-image generation through negation, counting, and spatial relations. The submultiplicative performance degradation highlights systemic limitations, showing that current architectures—optimized for visual plausibility—fundamentally mismatch the discrete logic required for compositional reasoning. Progress demands architectural innovations that integrate explicit symbolic mechanisms, training objectives that prioritize constraint satisfaction over perceptual quality, and evaluation metrics that measure true compositional understanding. Solving these core challenges is a prerequisite for trustworthy deployment in high-stakes applications and will enable models to reason about more abstract concepts like time and causality.

\section{Acknowledgments}
Vatsa and Singh are supported through Srijan: Center of Excellence on Generative AI.

\bibliography{aaai2026}

\end{document}